\documentclass[10pt,twocolumn,letterpaper]{article}

\usepackage{cvpr}
\usepackage{times}
\usepackage{epsfig}
\usepackage{graphicx}
\usepackage{amsmath}
\usepackage{amssymb}
\usepackage{comment}
\usepackage[title]{appendix}
\graphicspath{{./figures/}}

\usepackage[pagebackref=true,breaklinks=true,letterpaper=true,colorlinks,bookmarks=false]{hyperref}

\cvprfinalcopy 

\def\cvprPaperID{1978} 

\ifcvprfinal\fi
\begin{document}

\setlength{\abovedisplayskip}{0pt}
\setlength{\belowdisplayskip}{0pt}
\setlength{\abovedisplayshortskip}{0pt}
\setlength{\belowdisplayshortskip}{0pt}

\title{Synthesizing Images of Humans in Unseen Poses}
\author{Guha Balakrishnan\\
MIT\\
{\tt\small balakg@mit.edu}\\
\and
Amy Zhao\\
MIT\\
{\tt\small xamyzhao@mit.edu}
\and
Adrian V. Dalca\\
MIT and MGH\\
{\tt\small adalca@mit.edu}\\
\and
Fredo Durand\\
MIT\\
{\tt\small fredo@mit.edu}
\and
John Guttag\\
MIT\\
{\tt\small guttag@mit.edu}
}

\maketitle


\begin{abstract}
We address the computational problem of novel human pose synthesis. Given an image of a person
and a desired pose, we produce a depiction of that person in that pose, retaining the appearance of both the person and background. We present a modular generative neural network that synthesizes unseen poses using training pairs of images and poses taken from human action videos. Our network separates a scene into different body part and background layers, moves body parts to new locations and refines their appearances, and composites the new foreground with a hole-filled background. These subtasks, implemented with separate modules, are trained jointly using only a single target image as a supervised label. We use an adversarial discriminator to force our network to synthesize realistic details conditioned on pose. We demonstrate image synthesis results on three action classes: golf, yoga/workouts and tennis, and show that our method produces accurate results within action classes as well as across action classes. Given a sequence of desired poses, we also produce coherent videos of actions. 
\end{abstract}

\section{Introduction}
\begin{figure}
\begin{center}
\includegraphics[width=\linewidth]{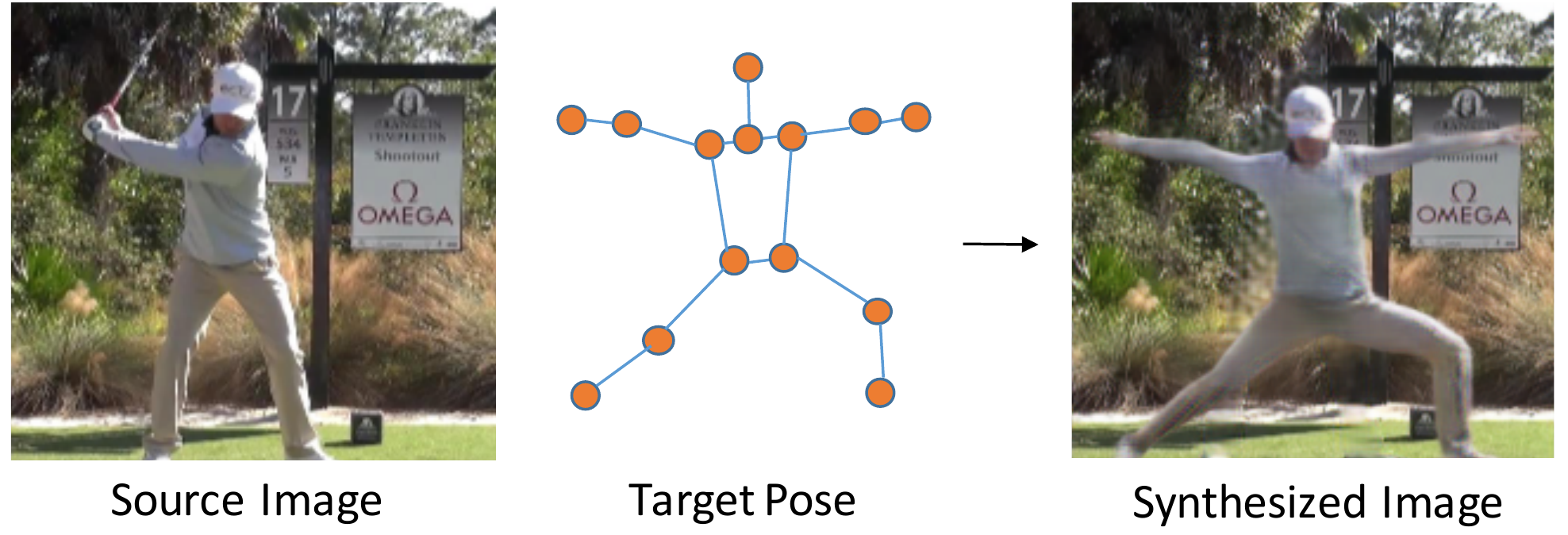}
\end{center}
\caption{Our method takes an input image along with a desired target pose, and automatically synthesizes a new image depicting the person in that pose. We retain the person's appearance as well as filling in appropriate background textures.}
\label{fig:teaser}
\end{figure}

Given an image of a person, we can imagine what that person would look like in a different pose. We are able to do this using a model of human appearance trained by observing many people in different contexts. In this work, we propose an automated method to address this task. Given an image of a person along with a target pose, we automatically synthesize a realistic image that depicts what the person would look like in that pose. We retain the appearance of both the person and the background in the transformation, as illustrated in Fig~\ref{fig:teaser}. 

To ensure a realistic image, we would like to retain the appearance of the person and background, and capture body part details consistent with the new pose. Differences in poses can cause complex changes in the image space, involving several moving parts and self-occlusions. Subtle details such as shading and edges should perceptually agree with the body's configuration. And background pixels that become disoccluded by the body must be filled in with appropriate content. 

We address these challenges by training a supervised learning model on pairs of images and their poses. Our model takes as input a source image and source 2D pose, and a desired 2D target pose, and synthesizes an output image. Our key idea is to decompose this complex problem into simpler, modular subtasks, trained jointly as one generative neural network. Our network first segments the source image into a background layer and multiple foreground layers corresponding to different body parts, allowing it to spatially move the body parts to target locations. The moved body parts are then modified and fused to synthesize a new foreground image, while the background is separately filled with appropriate texture to address gaps caused by disocclusions. Finally, the network composites the foreground and background to produce an output image. All of these operations are performed jointly as one network, and trained together using only a target image as a supervised label. 

Recent work has used images along with pose or viewpoint information to synthesize unseen views of people~\cite{lassner2017generative,ma2017pose,zhao2017multi}. These studies embed an input image and desired transformation into a latent space, and decode this space into an output image. In contrast, we decompose this complex problem into simpler subtasks, allowing us to generate more realistic results with limited data. Our layering approach decouples the foreground and background synthesis tasks, helping us synthesize better backgrounds. And by segmenting the foreground into body parts, we can model complex movements. The strategy of decomposing a network into modular subtasks has proven useful in recent learning models for visual reasoning~\cite{andreas2016learning,johnson2017inferring}. 

We demonstrate our model on images taken from 266 videos of three action classes downloaded from YouTube: golf, yoga/workouts, and tennis. Results show that our method can accurately reconstruct poses within a given action class, as well as transfer pose across action classes. Finally, by providing a sequence of poses and a source image, we show that we can construct a temporally coherent video portraying an action, despite our network not being explicitly trained to preserve temporal continuity. 

\section{Related work} 
View synthesis is a task in computer vision in which unseen camera views or poses of objects are synthesized given a prior image. Most view synthesis work has focused on simple rigid objects such as cars and furniture~\cite{ji2017,kulkarni2015,park2017transformation,rematas2016,yang2015,zhou2016view}. Recent studies have synthesized unseen views of people~\cite{lassner2017generative, ma2017pose,zhao2017multi}. These methods use encoder neural networks to capture complex relationships between the input image and desired transformation, and a decoder to synthesize the output image. In contrast, we represent a scene as separately manipulatable layers, allowing us to frame the task as a composition of simpler, modular subproblems. This allows us to explicitly move body parts and synthesize realistic backgrounds unlike the related studies. 

Many related problems in computer vision can be posed as an instance of image translation, or converting one representation of a scene into another. Examples include scene segmentation~\cite{quan2016,ronneberger2015}, surface normal prediction~\cite{eigen2015}, coloring~\cite{zhang2016}, style transfer~\cite{li2016,shih2013}, edge detection~\cite{xie2015} and sketch inversion~\cite{yagmur2016}. In these examples, pixels are modified rather than moved from input to output image. A recent study has shown that a UNet (an encoder-decoder neural net with skip connections) is capable of handling a wide variety of translation tasks~\cite{isola2016}. We incorporate the UNet architecture for several image translation subtasks in our problem, such as image segmentation and hole filling.

We use a GAN (generative adversarial network)~\cite{goodfellow2014,radford2015,salimans2016,zhao2016} to inject realism into our synthesized images. A GAN consists of a generator network that synthesizes candidates, and a discriminator network that classifies whether the candidate is real or synthesized. The generator's training objective is to increase the error rate of the discriminator, i.e. ``fool the discriminator,'' by producing instances that appear to be real. We use a conditional GAN, which synthesizes images conditioned on an input image and target pose. Previous works have conditioned GANs on images for various tasks such inpainting~\cite{pathak2016}, image prediction from a normal map~\cite{wang2016}, style transfer~\cite{li2016}, future frame prediction~\cite{mathieu2015}, and image manipulation~\cite{zhu2016}. An application-agnostic conditional GAN for image synthesis was proposed in ~\cite{isola2016}.



\begin{figure}[h!]
\begin{center}
\includegraphics[width=\linewidth]{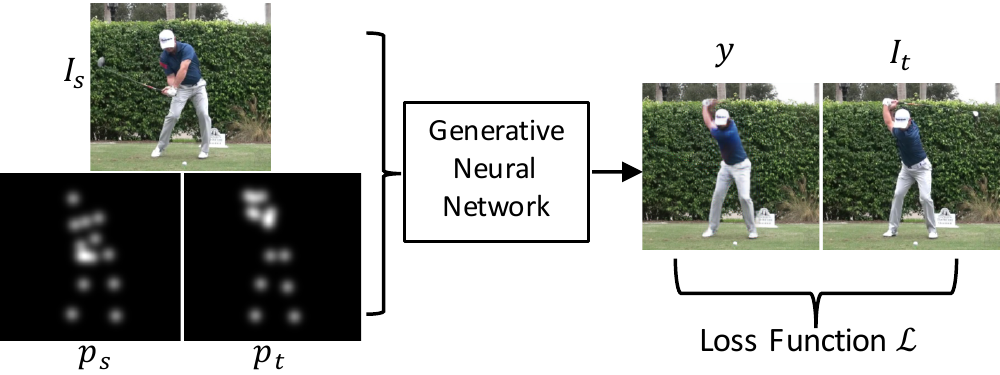}
\end{center}
\caption{Our network takes as input a tuple of the form ($I_{s}, p_{s}, p_{t}$), and synthesizes an image $y$. During training, a loss function $\mathcal{L}$ is used to minimize error between $y$ and $I_t$. We visualize $p_s$ and $p_t$ here as single-channel images, though in our model they contain a separate channel for each joint.}
\label{fig:example}
\end{figure}

\begin{figure*}[h!]
\begin{center}
\includegraphics[width=\textwidth]{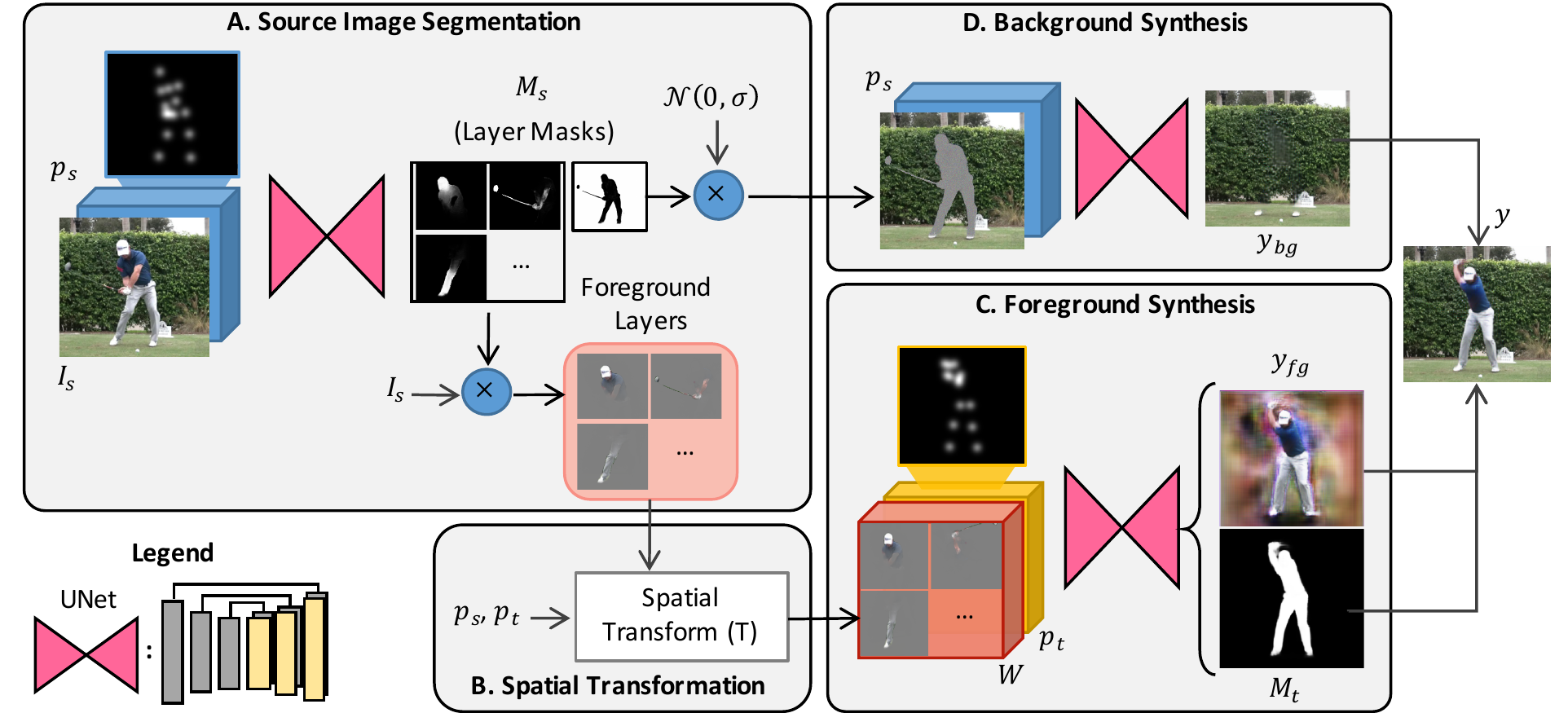}
\end{center}
\caption{Our network architecture, consisting of four modules. Module A performs image segmentation on $I_s$, separating the person's body and objects held by the person from the background. Module B spatially transforms the body parts in $I_s$. Module C synthesizes a target foreground image $y_{fg}$ by fusing the body parts in a realistic manner. This module also simultaneously outputs a foreground mask $M_t$. Module D synthesizes a background image, $y_{bg}$ via hole-filling. Finally, we composite $y_{fg}$ and $y_{bg}$ to produce $y$.}
\label{fig:network}
\end{figure*}

\section{Method}
We present a neural network model that learns to translate pose changes to the image space. Our model is trained on (example, label) tuples of the form (($I_{s}, p_{s}, p_{t}), I_t)$, where $I_{s}, p_s$ and $p_t$ are the source image, source 2D pose and target 2D pose, and $I_t$ is the target image (Fig.~\ref{fig:example}). We assume that $I_s$ and $I_t$ depict the same background and person in the same attire.

We design our network in a modular way to address several challenges. The motion field induced by a pose transformation often involves several moving body parts, large displacements, and occlusions. To address this, our model first segments the scene into foreground and background layers. It further segments the person's body into different part layers such as the arms and legs, allowing each part to then be moved independently of the others. Pixels in the background layer that become disoccluded are filled in with appropriate appearance. To render subtle shading and high-frequency details on the body, we use a combination of feature and adversarial losses that enforce realism of image details conditioned on the target pose. 

Fig.~\ref{fig:network} depicts our network, split into four modular subtasks. First, it separates the person's body parts from the background (module A: source segmentation). Next, it spatially moves the body parts to target locations (module B: spatial transformation). The network then fuses body parts into a coherent foreground (module C: foreground synthesis). Parts of the image disoccluded by the body are filled in with realistic texture (module D: background hole-filling). Finally, the foreground and background are composited to produce an output image $y$. We design our network such that these modules are learned jointly and trained using only the target image as a label.

Modules A,C and D are parametrized using separate UNet-style architectures~\cite{isola2016,ronneberger2015}. UNets have proven successful at image synthesis tasks where there is no movement between input and outputs. The only module of our model that does not use a UNet is the spatial transformer (module B), which is responsible for handling movements in the scene. Details of our three UNet architectures are found in supplementary material. We now describe our network in more detail. 

\subsection{Pose Representation}
As in past work on pose estimation~\cite{newell2016stacked,wei2016}, we represent the 2D poses $p_{s}$ and $p_{t}$ as 3D volumes in $\mathcal{R}^{H\times W\times J}$, where $H,W$ are the height and width of the input images and each of the $J$ channels contains a Gaussian ``bump'' centered at the $(x,y)$ location of a different joint. This representation allows the network to quickly leverage the spatial nature of the pose input in contrast to a flattened, dense representation. The spatial Gaussians also act as a regularization on the pose estimates which can be useful when joint locations are noisy. Joints outside the image domain are naturally represented with blank channels. In our experiments, we use the following $J=14$ joints: head, neck, shoulders, elbows, wrists, hips, knees and ankles.  

\begin{figure*}[h!]
\begin{center}
\includegraphics[width=\textwidth]{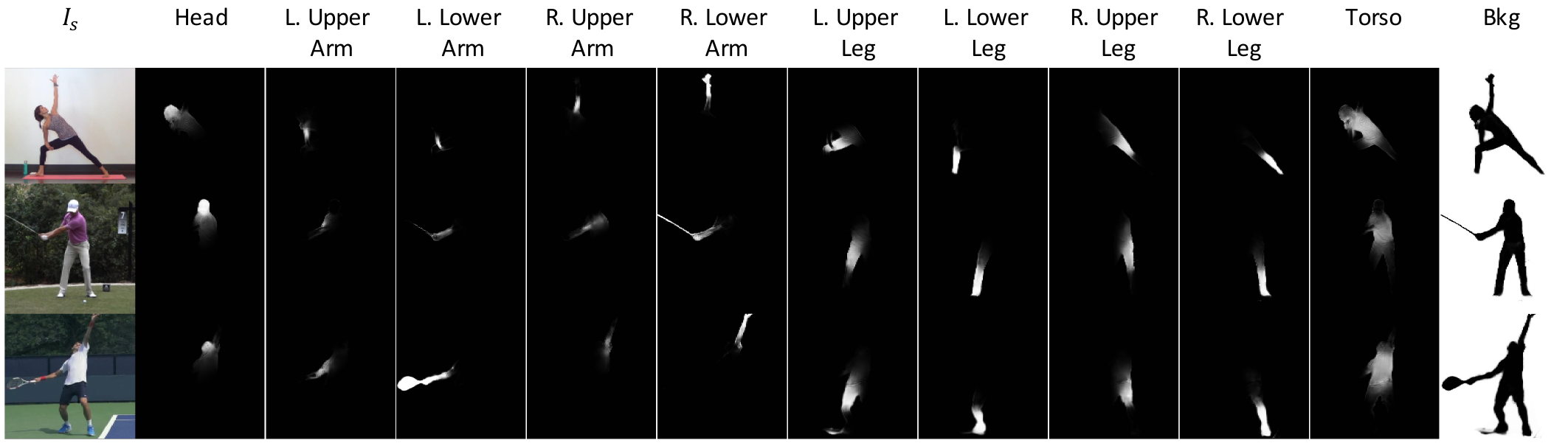}
\end{center}
\caption{Example outputs of the source image segmentation stage. Shown are the masks for each body part and background for three different examples. An interesting result is that commonly moving foreground structures like the golf club (2nd example), tennis racket (3rd example) and shadow (3rd example) are learned to be included with the foreground masks.}
\label{fig:segmentation_examples}
\end{figure*}

\subsection{Source Image Segmentation}
\label{sec:src_matting}
Moving scenes can be understood as a composition of layers~\cite{wang1994}. When a person moves, each body part may move differently from one another, typically leading to piecewise affine motion fields (not accounting for occlusions) in the image space. To handle such movement, we first segment $I_s$ into $L$ foreground layers and one background layer. The $L$ layers correspond to $L$ predefined body parts. We split the body into $L=10$ parts: head, upper arms, lower arms, upper legs, lower legs and torso. A body part is not the same as a joint; the first 9 parts consist of $2$ joints, and the torso contains $4$ joints. 

Our segmentation module is based on a UNet-style architecture (see Fig.~\ref{fig:network}). The input to the UNet is a concatenated volume $\left[I_s,p_s\right]\in \mathcal{R}^{H\times W\times (3+J)}$. The output of the UNet is a volume $\Delta M_s \in \mathcal{R}^{H\times W\times (L+1)}$. We add $\Delta M_s$ to an input (and therefore, unlearned) volume $\hat{M}_{s}$ specifying the rough location of each body part in $I_s$ to obtain our final mask volume $M_s = \text{softmax}\left(\Delta M_s + \log \hat{M}_{s}\right)$. $\hat{M}_{s}$ consists of a 2D Gaussian mask over the approximate spatial region of each body part, and helps our network converge to the segmentation we desire. The softmax function is used to enforce a probability distribution over layers. $\Delta M_s$ may be viewed as a residual component added to the coarse estimate $\hat{M}_{s}$.

Fig.~\ref{fig:segmentation_examples} shows sample masks produced by our method. We produce hard boundaries between the foreground and background, and soft boundaries between body parts. This is because neighboring body parts have correlated movement and appearance while the foreground and background do not. In our videos, humans are often holding small objects such as a golf club or tennis racket (as seen in examples 2-3 in Fig.~\ref{fig:segmentation_examples}). Our model learns to include body shadows and objects held in the hand with the foreground because they generally move with the body. 

We multiply each mask $M^l_s$ pixelwise across all three channels of $I_s$ (denoted by $\otimes$), to obtain masked images of each layer. Let $I^I_s = M^l_s \otimes I_s$ be the image that depicts layer $l$. We use $\{I^l_s\}_{l=1}^{L+1}$ in the subsequent modules of our model.

\subsection{Foreground Spatial Transformation}
The source segmentation stage separates the image into $L$ foreground layers, allowing us to move each layer separately from one another. Layers correspond to rigid body parts, which may be assumed to follow simple, parametric motions. We therefore apply a separate geometric transformation $T^l \in \mathcal{R}^{2\times 3}$ to each $\{I^l_s\}_{l=1}^{L}$. We compute $T^l$ using a similarity transformation fit using the joint positions of part $l$ in $p_s$ and $p_t$. Note that these transformations are \emph{not} learned, but are directly computed from the input poses. A similarity transformation accounts for translation, rotation and scaling of the body part. 

We warp $I^l_s$ using $T^l$ with a bilinear interpolation function \cite{jaderberg2015}, yielding a warped foreground image $W^l$. The bilinear interpolation function takes an input image along with a dense displacement field and outputs the image warped by the field. Critically, this layer has smooth gradients almost everywhere, and can be trained along with the rest of the network during backpropagation. Let the transform $T^l$ map pixel $(x,y)$ to subpixel location $(x',y')$. The bilinear interpolation layer computes $W^l$ with the following equation:

\footnotesize
\begin{equation}
W^l(x,y) = \sum_{q \in \mathcal{N}(x',y')}{I^l_s(q)(1-|x'-q_x|)(1-|y'-q_y|)}, 
\end{equation}
\normalsize

\noindent where $\mathcal{N}(x',y')$ is the set of four pixel neighbors of subpixel location $(x',y')$. Body part $l$ depicted in $W^l$ is now roughly at the correct location, scale and orientation for the target pose. Errors of subsequent layers are backpropagated through $I_s^l$, allowing $M_s^l$ to be learned (see Fig.~\ref{fig:segmentation_examples}).

\subsection{Foreground Synthesis}
The foreground synthesis branch (module $C$ in Fig.~\ref{fig:network}) merges the transformed body parts and further refines their appearance. We use a UNet that takes a concatenated volume $[W, p_t] \in \mathcal{R}^{H\times W\times (3L+J)}$ and outputs the target foreground $y_{fg}$ as well as a target mask $M_t\in \mathcal{R}^{H\times W\times 1}$. The two outputs are produced from the same UNet by branching off two separate convolutional layers at the end of the decoding stage. The target pose $p_t$ provides additional context as to where joints should be in the target image. 

\begin{figure}[h!]
\begin{center}
\includegraphics[width=\linewidth]{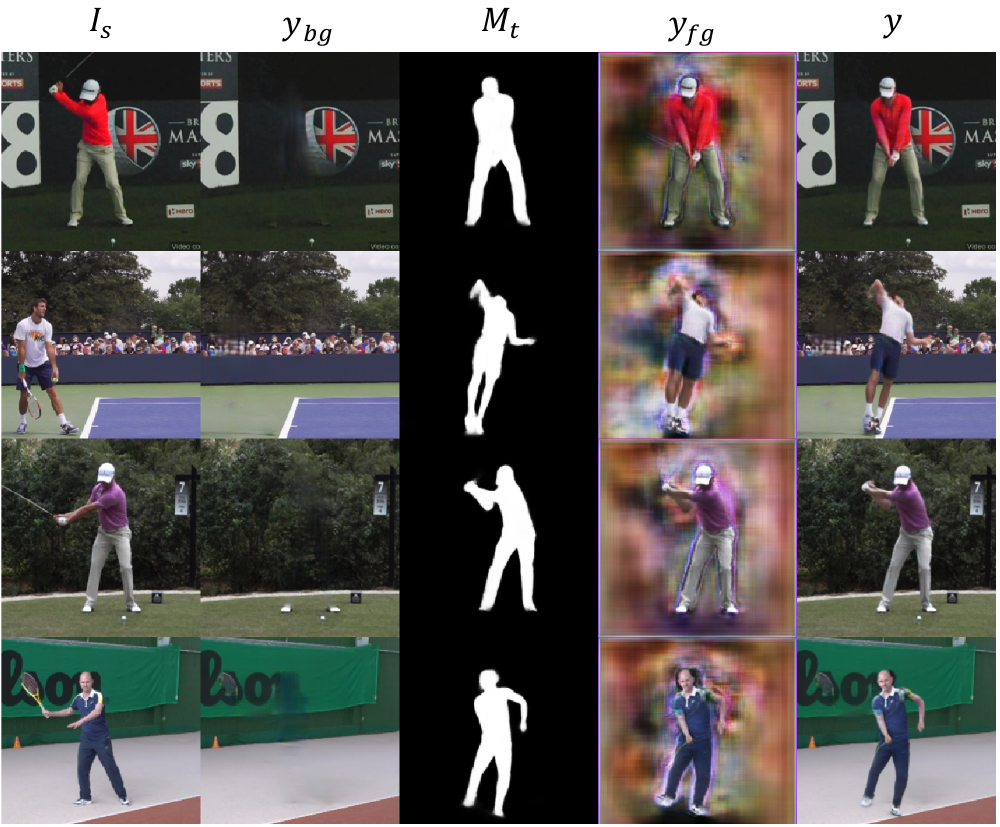}
\end{center}
\caption{Example outputs of the background (column 2) and foreground (columns 3-4) synthesis stages, as well as final outputs (column 5). Column 1 is the source image.}
\label{fig:fgbgmerge}
\end{figure}

Fig.~\ref{fig:fgbgmerge} shows several output examples of the mask $M_t$  (column 3) as well as $y_{fg}$ (column 4) for this stage. The body is realistically rendered even for dramatic pose changes. Objects being held by the person, such as the golf clubs in examples 1 and 3, or the tennis rackets in examples 2 and 4, are not retained, because they exhibit inconsistent patterns of movement given the body poses. Our model generates incoherent foreground pixels outside of the masked area (column 4) because these values do not affect our loss function. 

\subsection{Background Synthesis}
The background synthesis stage (module D in Fig.~\ref{fig:network}), fills in pixels of the scene occupied by the foreground in $I_s$. We provide image $I^{L+1}_s$ as input, which consists of the background pixels of $I_s$ and Gaussian noise in place of the foreground: $I^{L+1}_s = I_s \otimes M^{L+1}_s + \mathcal{N}(0,\sigma) \otimes (1 - M^{L+1}_s)$. Initializing the foreground with noise provides high-frequency gradients useful for texture synthesis~\cite{gatys2015}. We pass $[I^{L+1}_s, M^{L+1}_s,p_s]\in \mathcal{R}^{H\times W \times (4+J)}$ through a UNet to synthesize an image $y_{bg}$ with former foreground pixels assigned realistic values consistent with the background. We include $M^{L+1}_s$ and $p_s$ as inputs to provide additional context as to where the foreground is located in $I^{L+1}_s$.

See Fig.~\ref{fig:fgbgmerge}, column 2, for example background outputs of this stage. Our method is able to fill in backgrounds of varying colors and textures. Occasionally, parts of the person's feet are left in the background, such as the golf example in the third row. We hypothesize that this is because our dataset is biased towards golf videos with static feet. Objects being held by the person are also sometimes included in the background, such as a portion of the tennis racket in the fourth example.

\subsection{Foreground/Background Compositing}
We composite the target background and foreground images with a linear sum, weighted by the target mask $M_t$:
\begin{equation}
y = M_t \otimes y_{fg} + (1 - M_t) \otimes y_{bg}
\end{equation}
Column 5 of Fig.~\ref{fig:fgbgmerge} shows examples of our outputs. 

\subsection{Loss Functions}
\begin{figure}[h!]
\begin{center}
\includegraphics[scale=1]{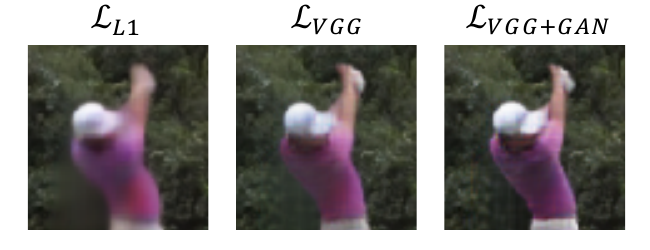}
\end{center}
\caption{Output image patches using different loss functions.}
\label{fig:loss_example}
\end{figure}

Let our generative model be denoted by function $G$: $y = G(I_s,p_s,p_t)$. A popular loss function for image synthesis is the L1 error between a synthesized and target image in RGB space: $\mathcal{L}_{L1}(G) = \mathbb{E}_{I_s,I_t,p_s,p_t}\left[\|y - I_t\|_1\right]$. Our experiments show that $\mathcal{L}_{L1}$ produces blurry images and does not prioritize high frequency details. The left-most image in Fig.~\ref{fig:loss_example} shows an example of our model's output using $\mathcal{L}_{L1}$. 

We therefore construct a feature loss $\mathcal{L}_{VGG}(G) = \mathbb{E}_{I_s,I_t,p_s,p_t}\left[\|\phi(y) - \phi(I_t)\|_1\right]$, which we compute by taking the L1 distance in a feature space $\phi$ constructed by concatenating activations from all channels of the first 16 layers of the VGG19 neural network pretrained for image classification~\cite{simonyan2014very}. VGG19 captures a range of image features from colors and edges in its initial layers, to textures, to common image structures in deeper layers. Minimizing error over all of these layers forces our network to capture various patterns. We normalize the activations of each channel by its mean and standard deviation in the training data. The center image in Fig.~\ref{fig:loss_example} shows an example of our model's output using $\mathcal{L}_{VGG}$.

While $\mathcal{L}_{VGG}$ is more effective than $\mathcal{L}_{L1}$ at synthesizing detail, it is not tailored to the distribution of images in our task. To address this we add an adversarial loss captured by a conditional discriminator network. The simultaneous use of a generator and adversary is known as a GAN. Our GAN consists of our generative network and an adversarial discriminator $D$ that outputs a probability that an image is real conditioned on the pose it should depict. $D$ will force $G$ to synthesize details consistent with the target pose. We define $L_{VGG+GAN}$ by:

\begin{equation}
\mathcal{L}_{VGG+GAN}(G,D) = \mathcal{L}_{VGG}(G) +  \lambda \mathcal{L}_{GAN}(G,D),
\end{equation} 
\normalsize

\noindent where $\mathcal{L}_{GAN}$ measures the binary cross-entropy classification error of the discriminator:

\begin{align}
\mathcal{L}_{GAN}(G,D) = &\mathbb{E}_{I_s,I_t,p_s,p_t}[\log D(I_t,p_t) +\nonumber \\
&\log (1-D(y,p_t))].
\end{align}
\normalsize

On its own, $\mathcal{L}_{GAN}$ will enforce that synthesized images lie within the distribution of all real images with target pose $p_t$. We add $\mathcal{L}_{VGG}$ to encourage $G$ to also approximate the true target image for each input. 

$D$ is trained simultaneously with $G$ on batches of synthesized images $y$ and corresponding targets $I_t$. In our experiments, we found $\lambda = 0.1$ to work well. The right-most image in Fig.~\ref{fig:loss_example} shows an example output using $\mathcal{L}_{VGG+GAN}$.
\begin{figure*}[t!]
\begin{center}
\includegraphics[width=\textwidth]{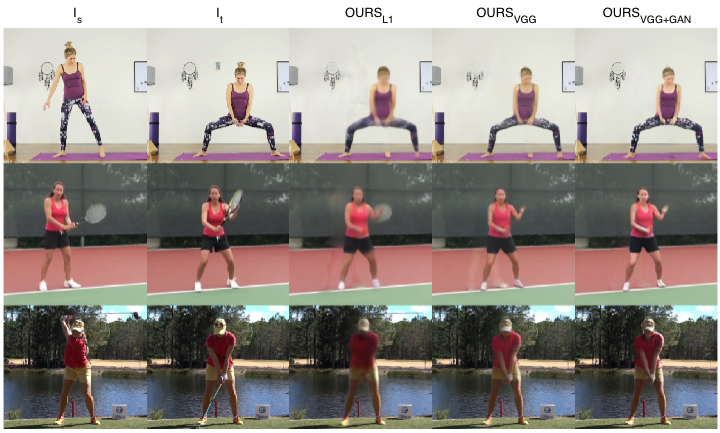}
\end{center}
\caption{Sample results of our method with different loss functions. Each row is a different example. L1 loss produces the blurriest images, while VGG+GAN produces the sharpest.}
\label{fig:results}
\end{figure*} 

\section{Experiments and Results}
We evaluate our method using videos of people performing actions collected from YouTube. Each training example is a pair of images and their corresponding poses taken from the same video. We select videos with mostly static backgrounds.  By using pairs of images from the same video, we hold a person's appearance and background constant while only allowing his/her pose to change. We collected videos from three action classes: golf swings, yoga/workout routines, and tennis actions. The dataset sizes are 136, 60 and 70 videos, respectively. We combine all action classes together into one dataset. We apply random data augmentations to each example: scaling, translation, rotation, horizontal flipping, and image saturation. We randomly held out $10\%$ of the videos for testing, and enforce that no individual appears in both the training and test set. We obtained 2D poses for video frames by first running an automated image pose estimator \cite{newell2016stacked} and then manually correcting dramatically incorrect joint positions.

We train separate networks using the three loss functions: $\mathcal{L}_{L1}$, $\mathcal{L}_{VGG}$ and $\mathcal{L}_{VGG+GAN}$. We use the ADAM optimizer and learning rate of $1e^{-4}$ when using $L_{VGG}$ or $L_{L1}$. All weights in these two networks are randomly initialized. We initialize the weights of the VGG+GAN network with the $L_{VGG}$ network, because we expect the discriminator to only adjust small details of the image. We implement our network in Keras with a Tensorflow backend.

In our first set of experiments, we use target poses extracted from frames that occur in the test videos as input, and compare the synthesized outputs to the known target frames. We also show that our network can synthesize realistic videos given a single image and a sequence of poses. In the second set of experiments, we transfer poses across action classes, e.g., synthesize a tennis player in a golfer's pose. 

\subsection{Within-Action Synthesis}
\begin{figure}[h!]
\begin{center}
\includegraphics[scale=0.6]{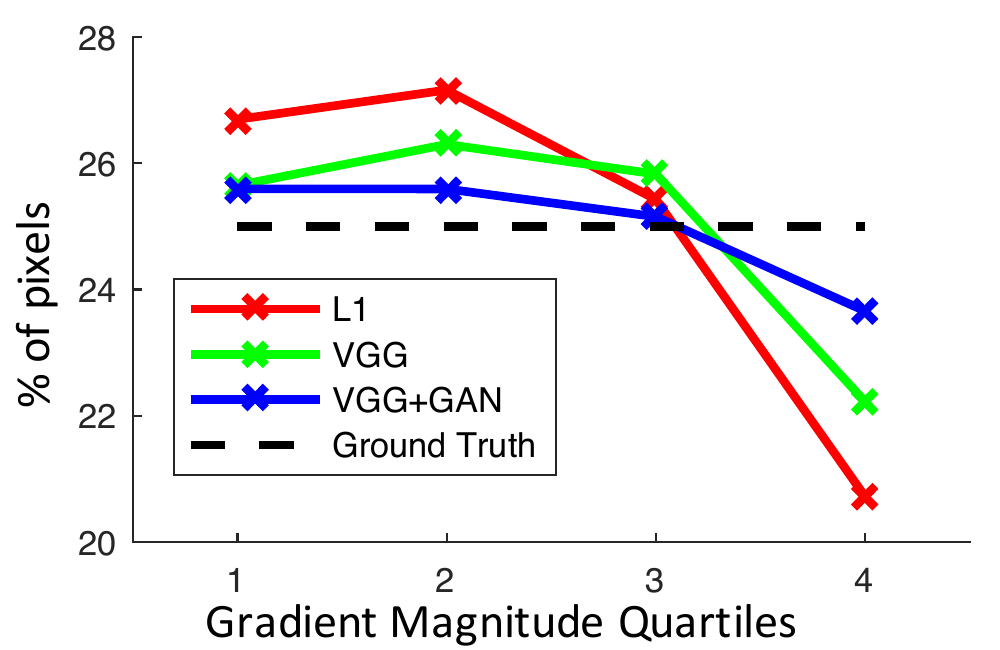}
\end{center}
\caption{Plot of pixel gradient magnitude for different loss functions, as well as for the ground truth images. We bin the gradients by ground truth quartiles. The VGG+GAN's distribution best matches the ground truth gradient magnitude distribution.}
\label{fig:gradient_comparison}
\end{figure}

Fig.~\ref{fig:results} presents our results using different loss functions. $\mathcal{L}_{VGG}$ synthesizes more details than $\mathcal{L}_{L1}$, and $\mathcal{L}_{VGG+GAN}$ further improves sharpness and realism. $\mathcal{L}_{VGG+GAN}$ is able to generate subtle details like the texture of the yoga instructor's pants (row 1), the lighting effects on the tennis player's body (row 2), and details of the golfer's arms and attire (row 3). We quantify the effect of our losses on image detail by plotting distributions of per-pixel spatial gradient magnitude in Fig.~\ref{fig:gradient_comparison}. We bin values into quartiles calculated from the gradient magnitude distribution of the real images. Among the losses, $\mathcal{L}_{L1}$ results in the highest percentage of small gradients, while $\mathcal{L}_{VGG+GAN}$ has the highest percentage of large gradients. VGG+GAN's distribution is closest to the real gradient magnitude distribution, plotted as the dashed black line.

\begin{figure}[h!]
\begin{center}
\includegraphics[width=\linewidth]{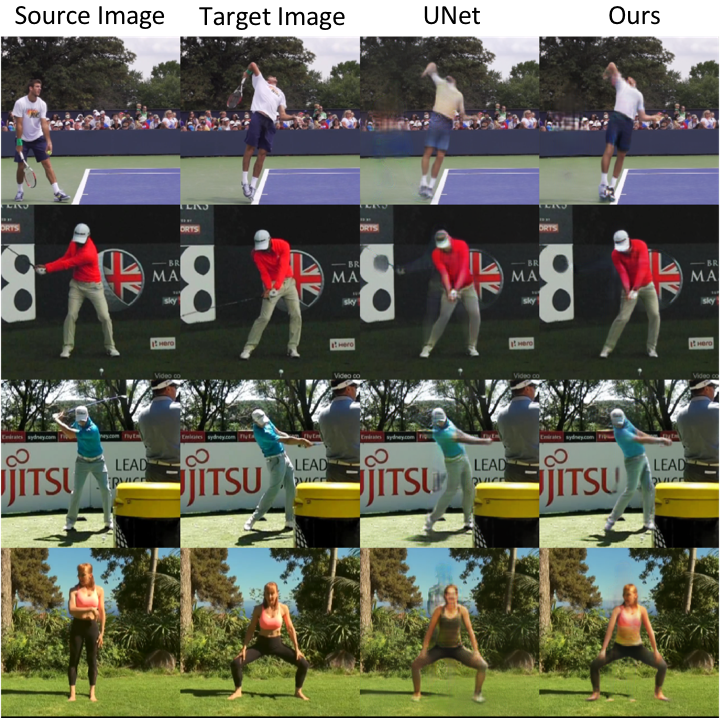}
\end{center}
\caption{A comparison of our model with a UNet when both are trained using $\mathcal{L}_{VGG+GAN}$. The UNet synthesizes incorrect foreground appearances in examples 1 and 4. It also produces background artifacts in all of these examples.}
\label{fig:unet_comparison}
\end{figure}  

As a baseline, we compare the accuracy of our method to a UNet architecture identical to our foreground synthesis network except that it takes an input volume consisting of $I_s$ (instead of $W$), $p_s$, and $p_t$. Variations of a UNet architecture have been used in a variety of image synthesis works, including some focusing on human synthesis~\cite{lassner2017generative, ma2017pose}. Fig.~\ref{fig:unet_comparison} compares our model with a UNet using $\mathcal{L}_{VGG+GAN}$. The UNet synthesizes incorrect foreground appearances as seen in examples 1 and 4. The UNet often copies appearances from similar poses in the training data rather than moving pixels. Our method does not suffer from this since we move body parts instead of synthesizing them from scratch. In addition, we are better at reconstructing backgrounds, such as the crowd in example 1, the black background in example 2, the ``U'' letter in example 3 and the sky in example 4. Table~\ref{tbl:losses} presents the performance of our approach vs. the UNet when training with $\mathcal{L}_{VGG+GAN}$. We evaluate performance using three metrics: L1 error, VGG error, and structural similarity score (SSIM)\cite{wang2004}.  SSIM is a common measure of perceived image quality, where a higher score is better. Our model outperforms the UNet for all metrics with statistical significance. 

\begin{table}[h!]
\centering
\caption{Errors (lower is better) and SSIM score (higher is better) of our method vs. a UNet architecture. Standard deviations are reported in parentheses. Our method achieves better scores. Differences are statistically significant using a paired t-test.}
\begin{tabular}{c c c c}
Model&L1 Error&VGG Error&SSIM Score\\
\hline
UNet&0.038(0.018)&0.215(0.091)&0.847(0.103)\\
Ours&0.034(0.018)&0.200(0.092)&0.863(0.105)\\
\end{tabular}
\label{tbl:losses}
\end{table} 

\begin{figure}[h!]
\begin{center}
\includegraphics[width=\linewidth]{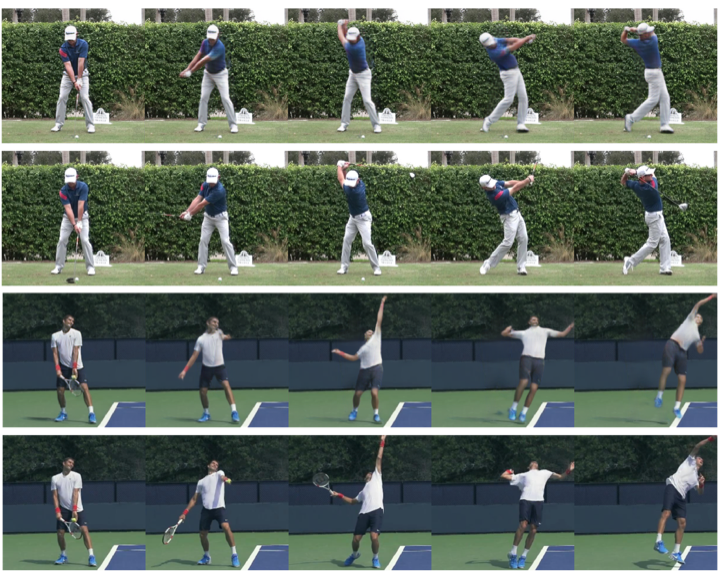}
\end{center}
\caption{Two example sequences generated by our method, along with ground truth sequences below them. We take the first frame of a real video (source image) along with the pose sequence of the entire video, and synthesize a sequence of images by independently applying our model to each target pose. Our method is able to produce temporally coherent appearances.}
\label{fig:video}
\end{figure}

\subsection{Video Synthesis}
\label{videosynthesis}
By applying our method independently to a sequence of target poses, we can construct a sequence of images depicting an action. Fig.~\ref{fig:video} shows examples of this for golf and tennis. We use the first frame of a video as the source image, and each pose of the sequence as a target pose. We show the ground truth video frames below our outputs for comparison. Our frames are temporally coherent, producing consistent appearances over time. The textures of the foreground, such as the creases on the golfer's pants and shadows on the tennis player's body are visually believable given the motion sequence, although they do not exactly match the ground truth. The synthesized background is consistent across frames because our model generates backgrounds using source image/pose information only. 

\subsection{Cross-Action Synthesis}
\begin{figure}[h!]
\begin{center}
\includegraphics[width=\linewidth]{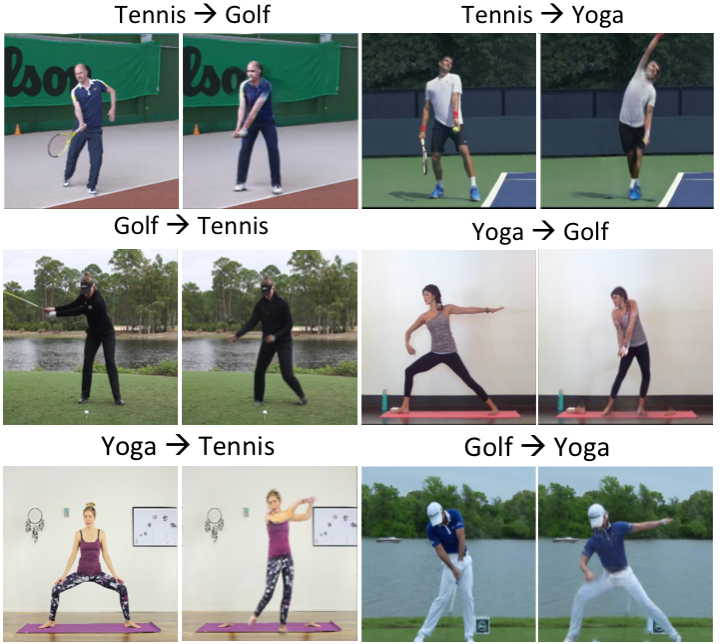}
\end{center}
\caption{Results when applying poses from one action class to images of another. We can successfully transfer poses across all classes.}
\label{fig:transfer}
\end{figure}

We now illustrate our network's ability to transform a person's pose to one of a different action class. We give a source image from one action class, and a target pose from a different class as input. Our network is identical to the one used in the last section, meaning that it has not seen examples of pose transformations across classes. Fig.~\ref{fig:transfer} presents some example outputs. We are able to produce realistic images across all action class permutations.

\section{Discussion}
Results show that our method is capable of synthesizing images across different action classes. Though trained on pairs of images within the same video, our model is able to generalize to pose-appearance combinations that it has never seen (e.g. a golfer in a tennis player's pose). This is possible because our model has learned to disentangle some pose and appearance characteristics. Our outputs are temporally consistent, producing coherent video sequences from a pose sequence. This is surprising since we apply our method independently to each frame and do not explicitly enforce temporal consistency. We believe this result can be attributed to how we move body layers to match the target pose instead of merging motion and appearance cues into a latent space as is common in past work.  

Our approach sometimes struggles with synthesizing detailed body parts like the face. Humans are particularly adept at detecting facial abnormalities, so improving the face is critical to synthesizing realistic images of people. To do this, a better metric is needed to assess image quality. Measures like L1 loss and SSIM provide some context but do not differentiate subtle differences, such as a deformed facial feature.

Transforming a 2D pose to an image is an underconstrained problem. For example, even if we know the 2D location of the wrist, the depth must be inferred. Furthermore, there could be many valid configurations of the body near that joint. Training using a limited set of action classes gives our network some contextual hints as to what configurations and appearances are likely given a particular 2D pose. This is exemplified by some of the cross-action transfers in Fig.~\ref{fig:transfer}. For instance, both the tennis instructor in row 1, column 2 and the yoga instructor in row 2, column 4 are synthesized with a golfer's glove after transforming their poses. As the number of possible poses that the network must resolve increases, it is likely that incorporating additional information like 3D joint locations will be desirable.

\section{Summary}
This work demonstrates the use of a modular generative neural network to synthesize images of humans in new poses. Our model performs synthesis in layers, decoupling the foreground from the background and different body parts from one another. It moves body parts to target locations, which allows it to capture large pose changes while maintaining correct foreground appearance. By decoupling the foreground from the background, it is also able to synthesize more realistic backgrounds than can a typical UNet architecture. Experiments also show that these design choices allow our model to generalize to tasks it was not explicitly trained for, such as transferring poses across action classes and producing temporally coherent action videos.

{\small
\bibliographystyle{ieee}
\bibliography{refs} 
}

\end{document}